\documentclass[journal]{IEEEtran}

\usepackage{amssymb}
\usepackage{bbm}
\usepackage{mathbbol}
\usepackage{newtxmath}
\usepackage{amsmath}
\usepackage{graphicx}
\usepackage{subfigure}
\usepackage{caption}
\usepackage{algorithm}
\usepackage{algorithmic}
\usepackage[top=0.85in, bottom=0.85in, left=0.75in, right=0.75in]{geometry}
\usepackage{bm}
\begin{document}
%
\title{VQ-DeepISC: Vector Quantized-Enabled Digital Semantic Communication with Channel Adaptive Image Transmission}

\author{Jianqiao Chen,
    Tingting Zhu, Huishi Song, Nan Ma,
    and Xiaodong Xu
\thanks{
Jianqiao Chen, TingTing Zhu and HuiShi Song are with the ZGC Institute of Ubiquitous-X Innovation and Applications, Beijing 100876, China.
Wenkai Liu, Nan Ma and Xiaodong Xu are with the State Key Laboratory of Networking and Switching Technology, Beijing University of Posts and Telecommunications, Beijing 100876, China. 
(e-mail: jqchen1988@163.com, zhutingting@zgc-xnet.com, songhuishi@zgc-xnet.com, manan@bupt.edu.cn, xuxiaodong@bupt.edu.cn)}
}

%


\maketitle
\thispagestyle{empty}
\begin{abstract}
Discretization of semantic features enables interoperability between semantic and digital communication systems, showing significant potential for practical applications. 
The fundamental difficulty in digitizing semantic features stems from the need to preserve continuity and context in inherently analog representations during their compression into discrete symbols  while ensuring robustness to channel degradation.
In this paper, we propose a vector quantized (VQ)-enabled digital semantic communication system with channel adaptive image transmission, named VQ-DeepISC. 
Guided by deep joint source-channel coding (DJSCC), we first design a Swin Transformer backbone for hierarchical semantic feature extraction, followed by VQ modules projecting features into discrete latent spaces. Consequently, it enables efficient index-based transmission instead of raw feature transmission. To further optimize this process, we develop an attention mechanism-driven channel adaptation module to dynamically optimize index transmission. 
Secondly, to counteract codebook collapse during training process, we impose a distributional regularization by minimizing the Kullback-Leibler divergence (KLD) between codeword usage frequencies and a uniform prior. Meanwhile, exponential moving average (EMA) is employed to stabilize training and ensure balanced feature coverage during codebook updates.
Finally, digital communication is implemented using quadrature phase shift keying (QPSK) modulation alongside orthogonal frequency division multiplexing (OFDM), adhering to the IEEE 802.11a standard. Experimental results demonstrate superior reconstruction fidelity of the proposed system over benchmark methods.
\end{abstract}


\IEEEpeerreviewmaketitle

\section{Introduction}

\IEEEPARstart{E}{mpowered} by the development of deep learning (DL) technology, semantic communication (SC)-grounded in understanding before transmitting, has been propelled to an emerging research topic in recent years [1]. By integrating deep joint source-channel coding (DJSCC), this approach achieves dual breakthroughs: near-optimal compression through structural semantic extraction, and exceptional resilience to dynamic channel variations [2]. Crucially, it prevents the abrupt quality deterioration known as \textit{cliff effect}, which is triggered once channel SNR breaches minimum operational requirements. Within this new paradigm of era of intelligent and context-aware communication, it has been demonstrated highly adaptable and extensible across diverse domains, including the text [3], images [4], [5], and video [6]. However, by mapping source data directly to channel symbols, most of recent researches place constellation points arbitrarily on the diagram, which deviate from the fundamental design principles underlying current digital communication systems. 

Achieving interoperability with digital communication systems, it requires semantic features to be translated into bits for transmission. Vector quantization (VQ)-based schemes, which convert features into indices and then bits for digital transmission, have recently gained widespread attention [7]-[10]. Authors in [8] propose a vector quantized variational autoencoder (VQ-VAE)-driven DJSCC scheme for wireless systems. By exploiting dataset-specific dependencies and channel state awareness, it develops optimized encoding/decoding architectures, which demonstrates VQ-VAE's inherent compatibility with digital communication systems. The image transmission system in [9] implements vector quantization within a U-Net framework, which extracts multi-scale semantic features while constructing hierarchical embedding spaces for feature quantization. However, it excessive storage requirements at both the transmitter and receiver ends. Moreover, the information bottleneck in the U-Net architecture restricts its capacity to model long-range contextual dependencies, consequently diminishing its effectiveness for high-resolution image processing. Built on VQ-VAE-2 [11], the post-deployment fine-tunable semantic communication (FTSC) framework proposed in [10] incorporates an updatable knowledge base that supports post-deployment fine-tuning, thus accommodating emergent image datasets during practical use. 

Overall, research on VQ-assisted digital semantic communication remains nascent. Several research aspects require further consideration. Given the inherent error-proneness of quantization process, the backbone network for semantic feature extraction requires meticulous architectural design, especially for high-resolution image processing. Concurrently, it is essential to protect the transmitted codebook indices to adapt to varying channel conditions. On the other hand, the non-differentiable nature of quantization layer inherently blocks gradient flow from the decoder back to the encoder within VQ architectures. Although straight-through estimator (STE) [7] offers a gradient approximation solution, this approach frequently results in suboptimal model performance and the codebook collapse phenomenon where a large percentage of the codebook converges to a zero norm and is unused by the model [12]. So it significantly constrains the information capacity of the VQ codebook. To fill these gaps, we propose a VQ-enabled digital semantic communication with adaptive image transmission, which is referred to as the VQ-enabled deep image semantic communication (VQ-DeepISC). The contributions of this paper are as follows:

\begin{itemize}
\item[]
\hspace{-0.5cm} $\bullet$
\hspace{-0cm}\textit{Multi-Stage Vector Quantization Framework for Semantic Feature Extraction:} 
We propose a cascaded architecture comprising: a Swin Transformer-based backbone network for hierarchical semantic feature extraction is designed, followed by which VQ modules projecting features into discrete latent spaces. Meanwhile, we develop an attention mechanism-driven channel adaptation module to dynamically optimize index transmission.

\hspace{-0.5cm} $\bullet$
\hspace{0.1cm}\textit{Kullback-Leibler Divergence (KLD)-Constrained Codebook Optimization:} 
To counteract codebook collapse, we impose a distributional regularization by minimizing the KLD between codeword usage frequencies and a uniform prior. Meanwhile, the exponential moving average (EMA) [7] is introduced to address the issues of training instability and uneven feature coverage during the codebook update process.
\end{itemize}

The rest of this paper is organized as follows. Section II provides overview of the system model. Section III elaborates the proposed VQ-enabled digital semantic communication system. 
Section IV presents experimental results that validate the performance of the proposed VQ-DeepISC system. Section V concludes the paper.

\textit{Notation:} Boldface small letters denote vectors and boldface capital letters denote matrices. 
$\mathbb{R}$, $\mathbb{Z}$ and $\mathbb{C}$ denote the real number field, the integer number field and the complex number field, respectively. 
$[\cdot]^{T}$ denotes matrix transpose. 
$\|\cdot\|_{2}$ denotes Euclidean norm.
$\ast$ denotes the convolution operation. 
$\mathcal{CN}(\cdot|\mu,\Gamma)$ denotes the complex Gaussian distribution with mean $\mu$ and covariance $\Gamma$.
$\mathbb{E}_{p(x)}[\cdot]$ denotes the expectation with respect to distribution $p(x)$.

\section{System model and problem formulation}

\begin{figure}[t]
\centering
\resizebox{0.5\textwidth}{0.18\textheight}{\includegraphics{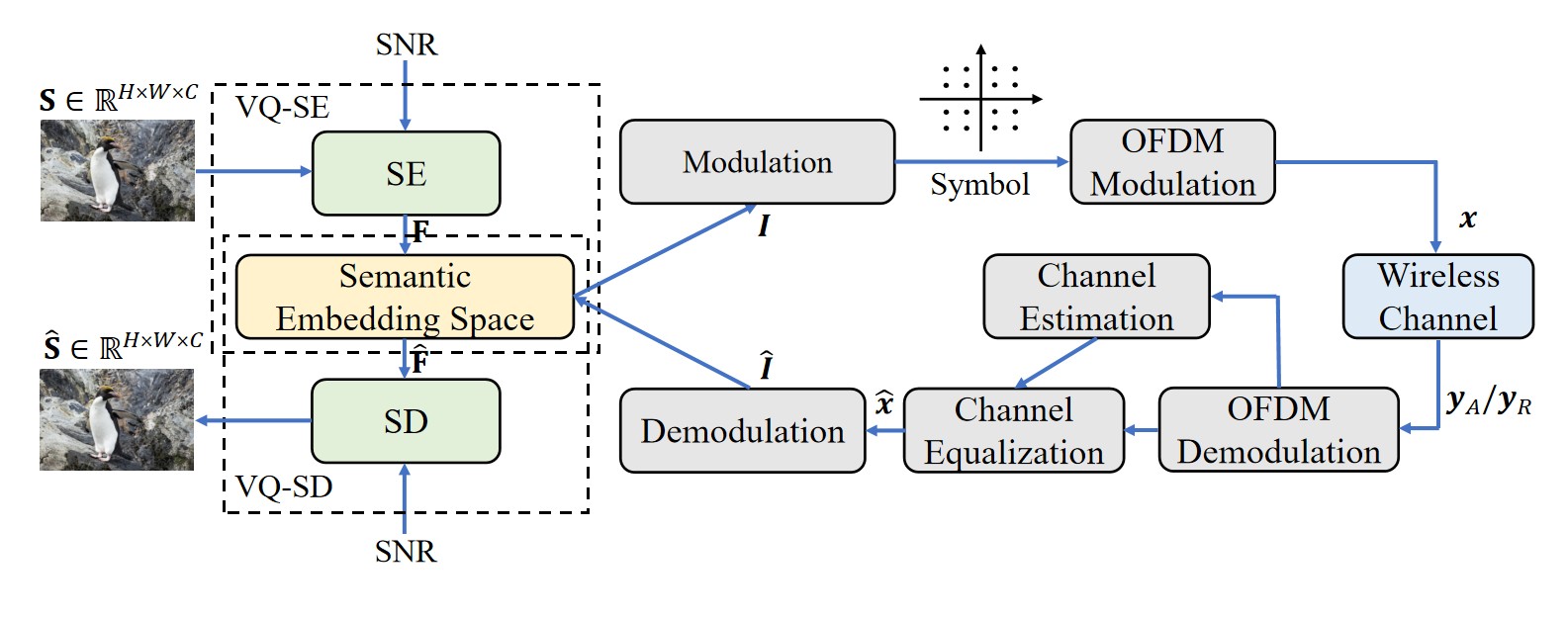}}
\captionsetup{font={footnotesize}}
\caption{The overall system architecture of the proposed VQ-DeepISC.}
\label{fig1}
\end{figure}


As shown in Fig. 1, the transmitter consists of three parts, namely the VQ-assisted semantic encoder (VQ-SE), Modulation, and OFDM Modulation. VQ-SE employs a shared quantized codebook to map semantic features to vectors in an embedding space, which ensures consistency across transmitter and receiver. The transmitter only transmits indices corresponding to semantic features rather than the features themselves, substantially enhancing efficiency. Meanwhile, robustness against dynamic channel conditions is ensured by embedding signal-to-noise ratio (SNR) directly into the model's training phase. Assume the input image of size $H$(height) $\times$ $W$(weight) $\times$ C(channel) as $\mathbf{\bm{S}} \in \mathbb{R}^{H\times W \times C}$. The semantic feature $\mathbf{\bm{F}} = \left[\bm{f}_{1}, \bm{f}_{2},..., \bm{f}_{M} \right]^{T} \in \mathbb{R}^{M \times K}$ of the input image can be expressed as 
\begin{equation}
\mathbf{\bm{F}} = SE_{\theta}\left(\mathbf{\bm{S}}, \mu \right),
\setcounter{equation}{1}
\end{equation}
where $SE_{\theta}(\cdot, \cdot)$ denotes the semantic extraction model with the learnable parameter set $\theta$, and $\mu$ denotes SNR that can be estimated at the joint source-channel decoder and fed back to the joint source-channel encoder. 

Assume that the local semantic embedding space shared by transceivers is denoted as $\mathbf{\bm{C}} = \left[\bm{c}_{1}, \bm{c}_{2},..., \bm{c}_{N} \right]^{T} \in \mathbb{R}^{N \times K}$, where $N$ represents the number of vectors, and $K$ represents the dimension of each vector. The VQ-SE is trained to map the high-dimensional input into a sequence of discrete latent vectors through minimizing Euclidean distance [8], i.e,
\begin{equation}
\bm{f}_{m} \leftarrow \bm{c}_{n},  \quad \mathrm{s.t}  \enspace n= \operatorname*{argmin}_{n} \left\| \bm{f}_{m} - \bm{c}_{n} \right\|_{2},
\setcounter{equation}{2}
\end{equation}
where $m\in [1, M]$ and $n\in [1, N]$. In the discretization process, we choose the nearest codeword $\bm{c}_{n}$ based on the closest distance from $\bm{f}_{m}$. Consequently, semantic features are converted into index sequences pointing to selected codewords, which is denoted as $\bm{I} \in \mathbb{Z}^{M \times 1}$. After modulation and OFDM modulation, $\bm{I}$ is processed as complex signal $\bm{x} \in \mathbb{C}^{B \times 1}$ for transmission, where $B$ is the length of the transmitted signal.

Signal transmission occurs over both additive white Gaussian noise (AWGN) and Rayleigh fading channels. As to AWGN channel, the channel output symbols $\bm{y} \in \mathbb{C}^{B \times 1}$ can be expressed as 
\begin{equation}
\bm{y}_{A} = \bm{x} + \bm{w},
\setcounter{equation}{3}
\end{equation}
where $\bm{w}\in \mathbb{C}^{B \times 1}$ satisfies independent and identically distributed (i.i.d) $\mathcal{CN}\left(0, \sigma^{2}\right)$, and $\sigma^{2}$ denotes noise power. 

As to Rayleigh channel, the channel output symbols can be expressed as
\begin{equation}
\bm{y}_{R} = \bm{h} \ast \bm{x} + \bm{w},
\setcounter{equation}{3}
\end{equation}
where $\bm{h}\in \mathbb{C}^{L \times 1}$ denotes Rayleigh channel with $L$ delays. 

After a series of subsequent signal processing steps at the receive side, namely the OFDM demodulation, channel estimation, equalization and demodulation, the recovered signal can be expressed as $\hat{\bm{x}} \in \mathbb{C}^{B \times 1}$. Using indices $\hat{\bm{I}}$ demodulated from $\hat{\bm{x}}$, the corresponding semantic embedding vectors are retrieved to construct feature tensor $\hat{\mathbf{\bm{F}}}$. Finally, the semantic decoder ultimately produces the reconstructed output as
\begin{equation}
\hat{\mathbf{S}} = SD_{\phi}\left(\hat{\mathbf{\bm{F}}}, \mu \right),
\setcounter{equation}{4}
\end{equation}
where $SD_{\phi}(\cdot, \cdot)$ denotes the semantic decoder model with the set of learnable parameters $\phi$.

Given SNR and semantic embedding constraints, we solve for encoder-decoder parameters $\theta^{*}$ and $\phi^{*}$ minimizing the expectation as follows:
\begin{equation}
\left(\theta^{*}, \phi^{*} \right) = \operatorname*{argmin}_{\theta, \phi} \mathbb{E}_{p(\mu)}\mathbb{E}_{p(\mathbf{S}, \hat{\mathbf{S}})}\left[d\left(\mathbf{S}, \hat{\mathbf{S}} \right)  \right],
\setcounter{equation}{5}
\end{equation}
where $\theta^{*}$ is the optimal parameter set of VQ-SE, $\phi^{*}$ is the optimal parameter set of VQ-SD, $p(\mu)$ denotes the probability distribution of SNR, $p(\mathbf{S}, \hat{\mathbf{S}})$ denotes the joint probability distribution of the original image $\mathbf{S}$ and the reconstructed image $\hat{\mathbf{S}}$, and $d\left(\mathbf{S}, \hat{\mathbf{S}} \right)$ denotes the distortion between $\mathbf{S}$ and $\hat{\mathbf{S}}$, namely $d\left(\mathbf{S}, \hat{\mathbf{S}} \right) = \frac{1}{n}\sum_{i=1}^{n}\left(s_{i} - \hat{s}_{i}\right)^{2}$, where $s_{i}$ and $\hat{s}_{i}$ represent the intensity of the color component of each pixel corresponding to $\mathbf{S}$ and $\hat{\mathbf{S}}$, respectively.

\section{VQ-enabled Digital Semantic Communication}
This section details our proposed VQ-enabled digital semantic communication system, which integrates semantic embedding space with dynamic channel adaptation. The DJSCC-based end-to-end architecture operates over stochastic physical channels using shared pre-trained semantic embeddings as vectorized priors.

\subsection{VQ-SE and VQ-SD for Semantic Extraction and Quantization}

\begin{figure}[t]
\centering
\resizebox{0.49\textwidth}{0.21\textheight}{\includegraphics{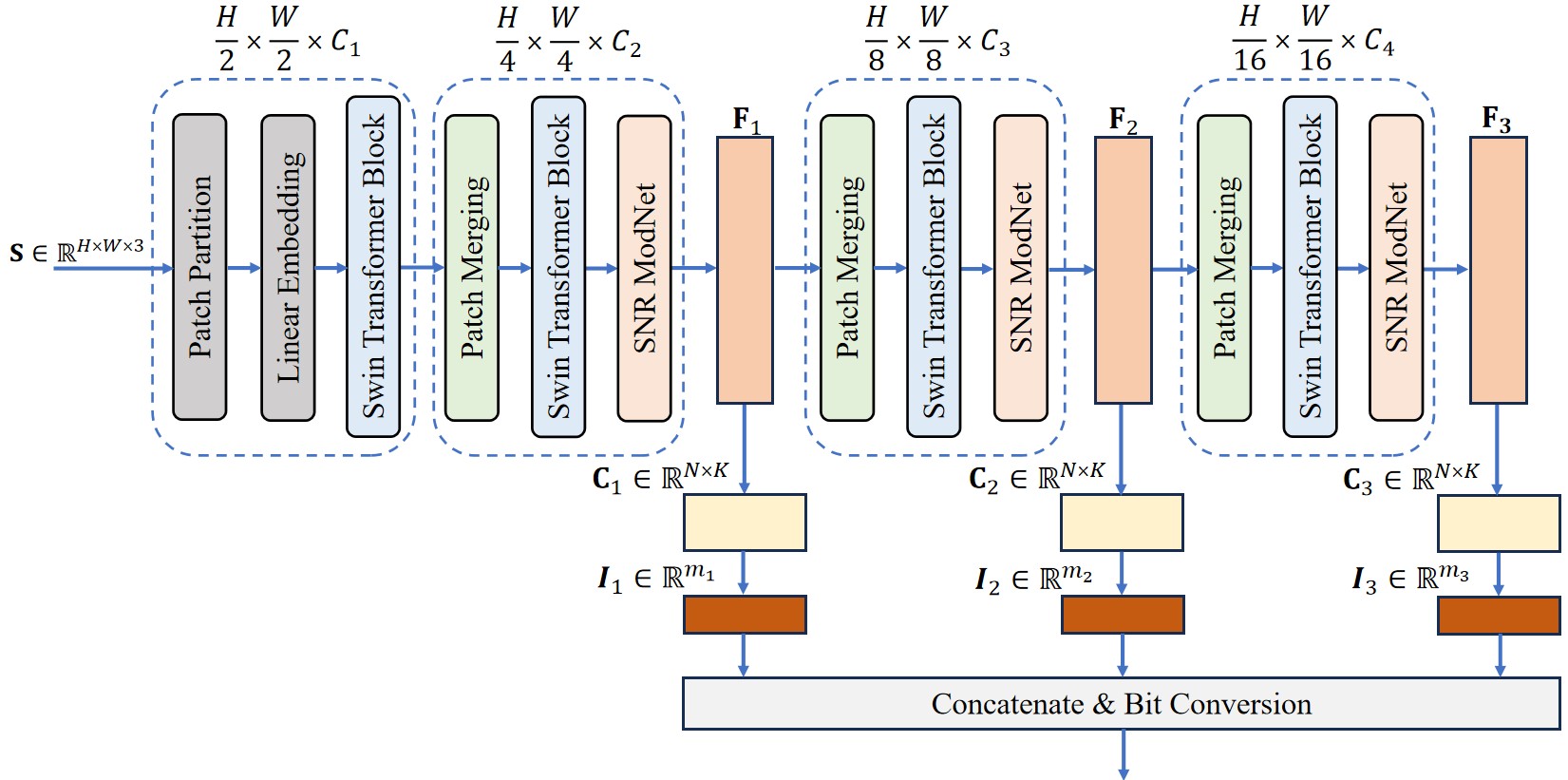}}
\captionsetup{font={footnotesize}}
\caption{The architecture of VQ-SE.}
\label{fig2}
\end{figure}

Fig. 2 illustrates the VQ-SE architecture for wireless image transmission, which integrates the Swin Transformer with VQ mechanism for joint semantic feature extraction and quantization. To mitigate quantization errors and enhance semantic representation, we design a multi-stage architecture where each stage aligns with distinct feature dimensions and dedicated quantized codebooks. The input RGB image source $\mathbf{S} \in \mathbb{R}^{H\times W \times 3}$ is first partitioned into $\frac{H}{2} \times \frac{W}{2}$ non-overlapping patches by \textit{Patch Partition} layer, which are regarded as tokens and arranged in a sequence by following a left-to-right and top-to-bottom order. Then it is followed by \textit{Linear Embedding} layer to map original feature into semantic latent representation with size $\frac{H}{2}\times \frac{W}{2} \times C_{1}$, where $1\times1$ convolution operation is adopted. Subsequently, a \textit{Swin Transformer Block} is applied to these features. 

The Swin Transformer employs two key attention mechanisms: window multi-head self-attention (W-MSA) and shifted window multi-head self-attention (SW-MSA). W-MSA restricts self-attention computation to non-overlapping local windows, dramatically reducing computational complexity compared to global attention. To overcome the limited cross-window connectivity, SW-MSA applies self-attention within windows shifted by half the window size relative to W-MSA. This shift enables communication between adjacent windows from the previous layer, which therefore allows the model to capture long-range dependencies. The size of these windows and the number of stacked Swin Transformer Blocks govern the extraction of semantic features at different scales. Details of W-MSA and SW-MSA design can be found in [13].

To build a hierarchical representation, each processing stage incorporates a \textit{Patch Merging} layer followed by \textit{Swin Transformer Blocks}. Concurrently, the \textit{SNR ModNet} is integrated to equip the model with adaptability across diverse channel conditions. For example, the neighboring embeddings are merged by a \textit{Patch Merging} layer, which results in the concatenated embeddings of size $C_{1}$ upgrading to size $C_{2}$. Subsequently, $\frac{H}{4} \times \frac{W}{4}$ patch embedding tokens with higher-resolution are fed into \textit{Swin Transformer Block}. Consequently, the architecture substantially increases the model capability by effectively capturing long-range interactions, leveraging global context, and processing complex high-resolution patterns efficiently. Since the indices are easily affected by different channel conditions, the \textit{SNR ModNet} for channel adaption is specifically designed, whose detailed architecture is presented in Section II-B. 

Assume $\mathbf{F}_{1}$, $\mathbf{F}_{2}$ and $\mathbf{F}_{3}$ as extracted features with different stages. Based on Eq. (2),  $\mathbf{F}_{1}$, $\mathbf{F}_{2}$ and $\mathbf{F}_{3}$ are mapped to the quantized codebooks  $\mathbf{C}_{1}$, $\mathbf{C}_{2}$ and $\mathbf{C}_{3}$, respectively. Then, we obtain the corresponding indices of the quantization vectors in codebooks, namely $\bm{I}_{1} \in \mathbb{R}^{M_{1}}$, $\bm{I}_{2}\in \mathbb{R}^{M_{2}}$ and $\bm{I}_{3}\in \mathbb{R}^{M_{3}}$. Finally, by using \textit{Concatenate} $\&$ \textit{Bit Conversion} layer, the indices are cascaded to form a vector, and then mapped into a bit sequence for transmission. Given the critical impact of quantized codebook expressiveness on semantic communication fidelity, 
Section II-C details our proposed method for counteracting codebook collapse.

\begin{figure}[t]
\centering
\resizebox{0.49\textwidth}{0.22\textheight}{\includegraphics{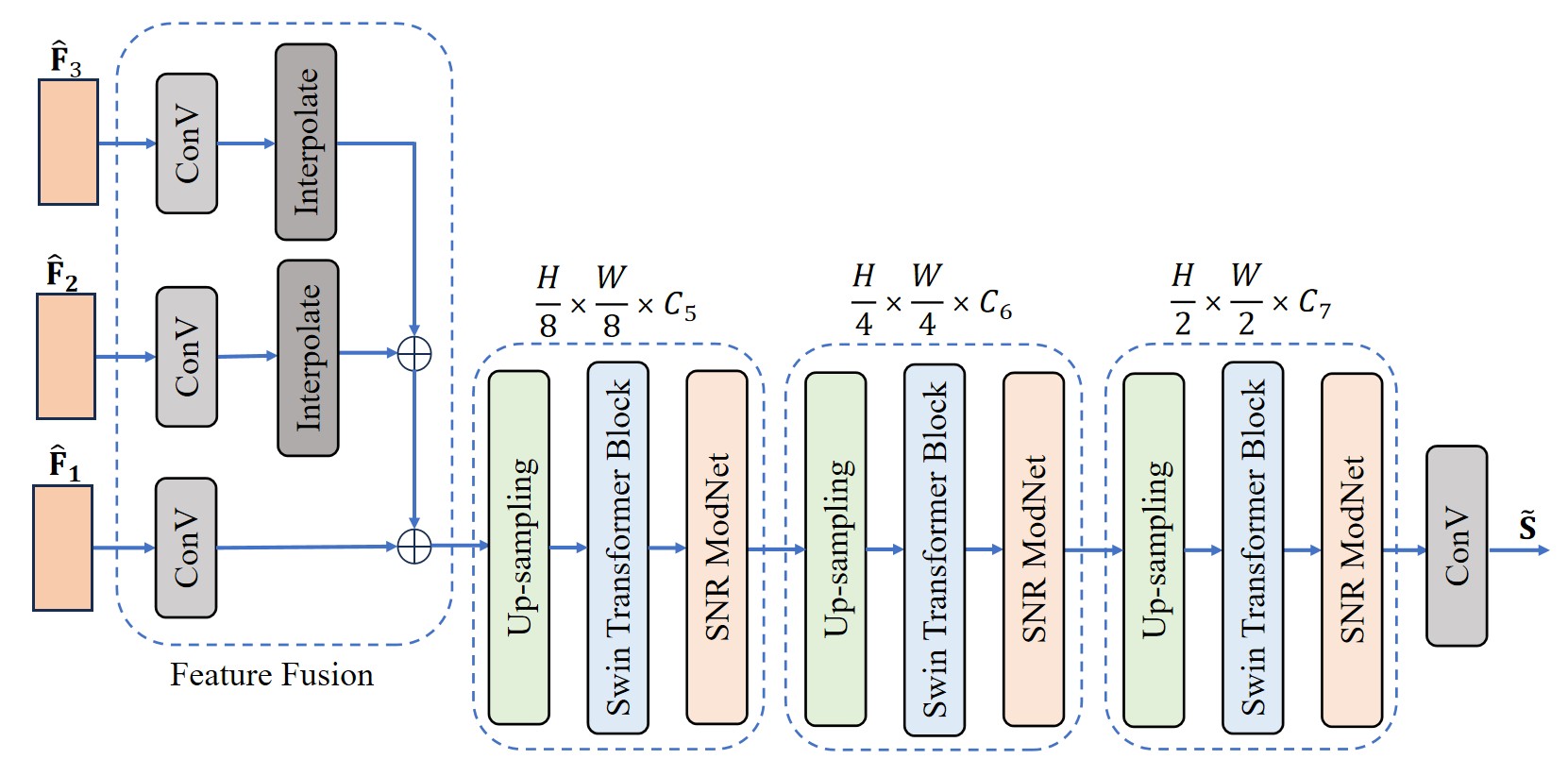}}
\captionsetup{font={footnotesize}}
\caption{The architecture of VQ-SD.}
\label{fig3}
\end{figure}

Fig. 3 illustrates the VQ-SD architecture for wireless image reconstruction, which incorporates a \textit{Feature Fusion} module and cascaded Swin Transformer-based reconstruction blocks. The  \textit{Feature Fusion} module first aligns multi-scale feature $\hat{\mathbf{F}}_{1}$, $\hat{\mathbf{F}}_{2}$, and $\hat{\mathbf{F}}_{3}$ via convolutional transformation \textit{ConV} and \textit{Interpolation}, and then synthesizes them through element-wise summation. The \textit{Up-sampling} layer, \textit{Swin Transformer Block} and \textit{SNR ModNet} are cascaded to gradually extract semantic features as well as adapt to channel conditions. For example, the features are up-sampled by a \textit{Up-sampling} layer, which results in the concatenated embedding of size $C_{5}$ reducing to size $C_{6}$. Subsequently, $\frac{H}{8}\times \frac{W}{8}$ patch embedding tokens with lower-resolution are fed into \textit{Swin Transformer Block}, and then a \textit{SNR ModNet} is followed to improve its adaptive ability to changing channel conditions. Note that the \textit{Up-sampling} layer is implemented using a two-dimensional transposed convolution operation, where kernel size and stride configurations control spatial resolution expansion. Finally, through the convolutional operation \textit{ConV}, the output that matches with size of source image is achieved. 

\subsection{SNR ModNet for Channel Adaption}

While sensitivity to channel changes remains a challenge for end-to-end semantic communication systems, research demonstrates that attention mechanisms can provide adaptive capabilities under varying channel conditions [14].
Taking advantage of this insight, we develop \textit{SNR ModNet}, which dynamically generates scaling parameters conditioned on the instantaneous SNR. 
Fig. 4 diagrams \textit{SNR ModNet}'s adaptive architecture leveraging channel-wise soft attention. Firstly, the SNR is input into \textit{SNR Extraction} module to balance the wide SNR range. Specifically, the \textit{SNR Extraction} module consists of \textit{Linear} operation and \textit{ReLU} operation. Secondly, the feature is processed by using a global average pooling function, denoted as \textit{Pooling} layer, which is followed by \textit{Factor Prediction} module to predict the scaling factor based on concatenated feature and SNR information. The \textit{Factor Prediction} module consists of \textit{Linear} operation and \textit{Sigmoid} operation. In addition, the \textit{Feature Enhancer} module is adopted as residual link for feature enhancement, which consists of \textit{ConV} operation and \textit{ReLU} operation. Finally, the features that have been superimposed with SNR information is obtained, and its performance analysis is detailed in Section IV.
\begin{figure}[t]
\centering
\resizebox{0.4\textwidth}{0.21\textheight}{\includegraphics{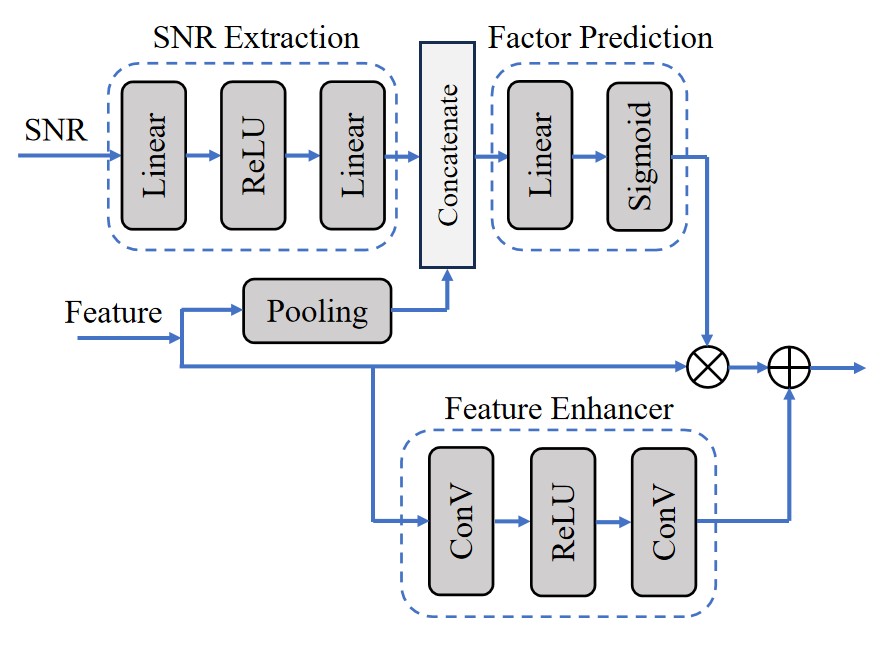}}
\captionsetup{font={footnotesize}}
\caption{The architecture of SNR ModNet.}
\label{fig4}
\end{figure}

\subsection{Codebook Updating and Loss Function}

Joint optimization of the quantization codebook with semantic encoder/decoder requires addressing two fundamental challenges. Firstly, the nearest-neighbor mapping operation, which assigns features to codebook embeddings via minimum distance criteria, induces gradient discontinuity. So, it prevents direct backpropagation to codebook parameters, which renders traditional gradient descent ineffective. Secondly, the insufficient representational capacity of codebook results in huge quantization error as to semantic encoding and decoding. In this case, some embedded vectors may rarely be utilized due to poor initialization or insufficient training, which therefore results in a waste of codebook capacity.

For the first issue, the STE is always adopted. For the second issue, we devise a novel scheme enforcing uniform embedding utilization across the codebook space. The core idea is to introduce the term based on KLD to quantify the discrepancy between the codebook usage and a uniform prior. Meanwhile, the EMA is adopted to stabilize training and ensure comprehensive coverage of feature space during codebook update.  The optimization procedure works as follows. 

In terms of the $e$-th training epoch, the cumulative number of semantic features that is mapped to the quantized vector $\bm{c}_{n}, n=1,2,...,N$, in codebook can be calculated as 
\begin{equation}
\varphi_{n}^{(e)} = \gamma\varphi_{n}^{(e-1)} + (1-\gamma)\psi_{n}^{(e)}, 
\setcounter{equation}{6}
\end{equation}
where $\varphi_{n}^{(e)}$ denotes the number of semantic features that are mapped to $\bm{c}_{n}$ in the $e$-th training epoch, $\psi_{n}^{(e)}$ denotes the current frequency usage of $\bm{c}_{n}$ in codebook, and $\gamma \in[0, 1)$ denotes the decay rate.

Assume that $p_{c}$ denotes the probability of using the codebook related to $\varphi_{n}^{(e)}, n=1,2,...,N$ and $p_{u}$ denotes the probability of uniform distribution. So, the KLD of $p_{c}$ and $p_{u}$ can be calculated as 
\begin{equation}
D_{KL}\left(p_{c} || p_{u} \right) = -H\left(p_{c} \right) + \log(N),
\setcounter{equation}{7}
\end{equation}
where $H(p_{c})$ denotes the entropy of $p_{c}$. In this case, $D_{KL}\left(p_{c} || p_{u} \right)$ is denoted as the distribution loss.

The cumulative sum of features that are assigned to the quantized vector $\bm{c}_{n}$ can be calculated as 
\begin{equation}
{\Phi}_{n}^{(e)} = \gamma\Phi_{n}^{(e-1)} + (1-\gamma)\sum_{\bm{c}_{n} \leftarrow \bm{f}_{m}}\bm{f}_{m}.
\setcounter{equation}{8}
\end{equation}
The codebook is updated as 
\begin{equation}
\bm{c}_{n}^{e} = \frac{\Phi_{n}^{(e)}}{\varphi_{n}^{(e)} + \epsilon}, n=1,2,...,N,
\setcounter{equation}{9}
\end{equation}
where $\epsilon$ denotes a hyperparameter.

So the loss function related with training quantized codebook consists of codebook loss, commitment loss and distribution loss. It can be expressed as 
\begin{equation}
\mathcal{L}_{VQ} = \left\| \mathrm{sg}[\mathbf{F}] - \mathbf{C} \right\|_{2}^{2} + \alpha \left\| \mathbf{F} - \mathrm{sg}[\mathbf{C}] \right\|_{2}^{2} + \beta D_{KL}\left(p_{c} || p_{u} \right), 
\setcounter{equation}{10}
\end{equation}
where $\mathrm{sg}[\cdot]$ denotes the stop-gradient operator that is defined as identity at forward computation time and has zero partial derivatives, and $\alpha$ and $\beta$ are hyperparameters.

Overall, the total loss function of whole digital semantic communication system can be expressed as 
\begin{equation}
\mathcal{L}\left(\mathbf{S}, \hat{\mathbf{S}} \right) = \left\|\mathbf{S} - \hat{\mathbf{S}} \right\|_{2}^{2} + \sum_{i=1,2,3}\mathcal{L}_{VQ}^{(i)}.
\setcounter{equation}{11}
\end{equation}
Note that the three-stage architecture of VQ-SE is intentionally structured to align distinct feature dimensions and quantized codebooks.

\section{Simulation Results and Discussions}

This section evaluates VQ-DeepISC under codebook optimization and channel variation. Without loss of generality, we restrict analysis to AWGN channels since the Rayleigh channel can be made equivalent through channel equalization. 
Models are trained on ImageNet [15] at 256×256 resolution. The best-performing model is then evaluated on DIV2K [16] using 1024×1024 resolution. Reconstruction fidelity is quantified to measure divergence between original and reconstructed images, including average peak signal-to-noise ratio (PSNR) for pixel-level accuracy and multi-scale structural similarity index (MS-SSIM) for perceptual quality assessment. 
Bit compression ratio (BCR) is defined as the ratio of the number of bits required for the final transmission sequence to the original images, which is given by $\mathrm{BCR} \triangleq \frac{B_{s}}{H \times W \times C \times 8}$, where $B_{s}$ denotes the image size in bits after compression.

The proposed model is implemented in PyTorch, and is optimized via AdamW optimizer with a learning rate of $2\times 10^{-4}$ and weight decay of $10^{-4}$. 
A cosine annealing scheduler is applied with a maximum iteration count $300$ and a minimum learning rate $10^{-6}$. The periodic modulation navigates saddle points while avoiding suboptimal convergence. Gradient clipping is implemented via setting a maximum norm of $5$ to prevent exploding gradients and ensure robust training. All of our experiments are performed on a Linux server with GTX 4090Ti GPU. During training process, other hyperparameters are set to be $\gamma=0.99$, $\epsilon=10^{-5}$, $\alpha=0.25$, and $\beta=0.05$.

\begin{figure}
\centering
\subfigure[]{
  \label{fig:subfig:a}
   \includegraphics[width = 0.47\textwidth]{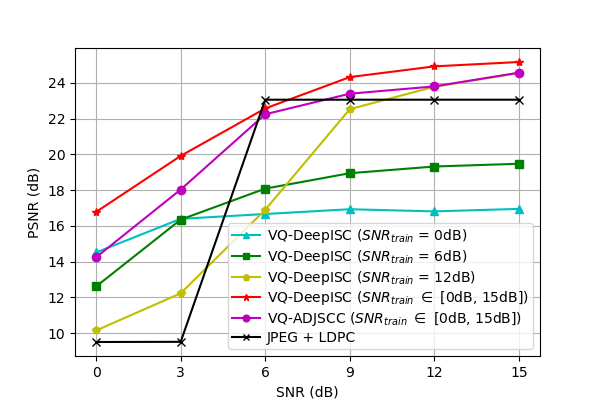}}
\hspace{1in}
\subfigure []{
 \label{fig:subfig:b}
 \includegraphics[width = 0.47\textwidth]{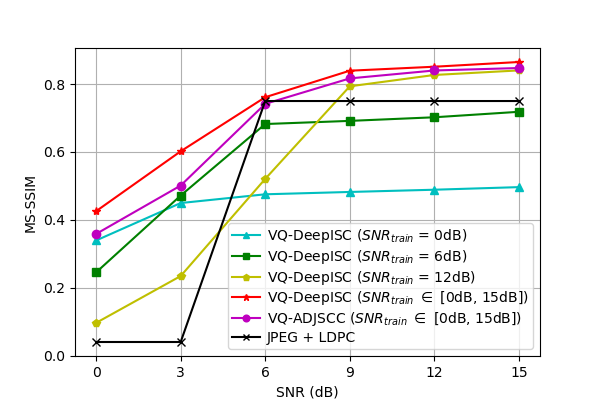}}
\captionsetup{font={footnotesize}}
\caption{Channel adaptive performance of VQ-DeepISC and counterparts on DIV2K test images. (a) PSNR vs. SNR and (b) MS-SSIM vs. SNR.}
\label{fig:subfig}
\end{figure}

The VQ-DeepISC model integrating  \textit{SNR ModNet} is trained on signals with SNR uniformly sampled from range [0dB, 15dB]. 
For benchmarking purposes, we employ three standalone VQ-DeepISC models lacking \textit{SNR ModNet}, each trained specifically at one fixed SNR level (0dB, 6dB, or 12dB).
Additionally, the traditional image compression model and ADJSCC model are chosen. The traditional model adopts the joint photographic experts group (JPEG) and low density parity check (LDPC) as the source encoding and channel encoding, respectively. To enable compatibility with digital semantic systems, we redesigned ADJSCC by fusing its SNR-adaptive module with our VQ-DeepISC framework (shown in Section II-A) for joint semantic extraction and quantization, which is referred to as VQ-ADJSCC. Digital communication is implemented using QPSK modulation alongside OFDM, adhering to the IEEE 802.11a standard.

Fig. 5 shows the PSNR and MS-SSIM comparison for VQ-DeepISC, VQ-ADJSCC, and BPG method with different compression ratio of source over AWGN channels. Specifically, the BCR of traditional method is 0.1, and BCR of other DL-based methods is 0.02. 
It can be observed that: 
1) VQ-DeepISC outperforms traditional methods in PSNR and MS-SSIM, particularly at medium-low SNRs, and shows smooth degradation without \textit{cliff effect}.
2) VQ-DeepISC surpasses VQ-ADJSCC in performance by leveraging the proposed SNR ModNet.
3) Higher $\mathrm{SNR_{test}}$ improves VQ-DeepISC's performance monotonically. Crucially, it maintains significant advantages over the SNR ModNet-free baseline regardless of deviations between $\mathrm{SNR_{train}}$ and $\mathrm{SNR_{test}}$.

\begin{figure}
\centering
\subfigure[]{
  \label{fig:subfig:a}
   \includegraphics[width = 0.47\textwidth]{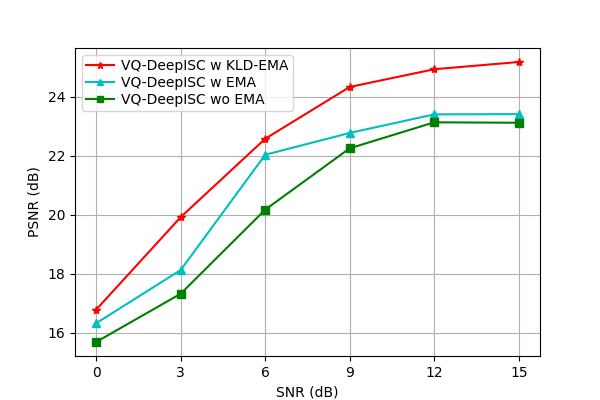}}
\hspace{1in}
\subfigure []{
 \label{fig:subfig:b}
 \includegraphics[width = 0.47\textwidth]{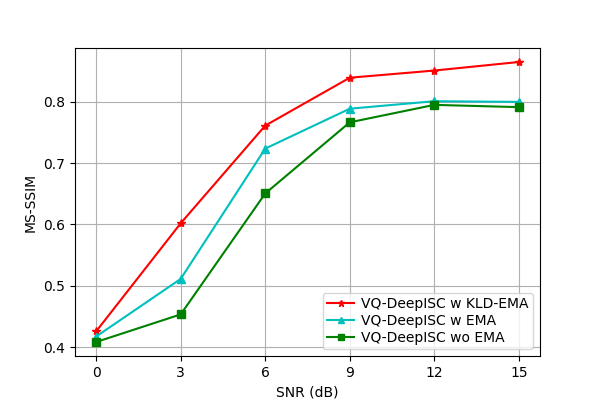}}
\captionsetup{font={footnotesize}}
\caption{Performance of VQ-DeepISC under different methods of updating codebook (a) PSNR vs. SNR and (b) MS-SSIM vs. SNR.}
\label{fig:subfig}
\end{figure}

The ablation experiment on the effectiveness of the codebook update method is presented. Fig. 6 shows the PSNR and MS-SSIM comparison for VQ-DeepISC model with considering different methods of updating codebook. Specifically, the VQ-DeepISC model with KLD-EMA, VQ-DeepISC model with EMA, and VQ-DeepISC model without EMA are simulated. 
Results confirm KLD-EMA achieves optimal PSNR/MS-SSIM, which validate the proposed codebook update strategy.

\section{Conclusion}
This paper introduces VQ-DeepISC, a pioneering vector-quantized digital semantic communication system optimized for adaptive image transmission over wireless channels. Within the DJSCC framework, our system develops a hierarchical Swin Transformer backbone for multi-scale semantic feature extraction. Subsequent VQ modules project features into discrete latent spaces, which enables integer-based index transmission instead of analog features. Meanwhile, an attention-driven adaptation module dynamically optimizes transmission by adjusting encoding to instantaneous channel conditions. 
To address codebook learning challenge, we propose a KLD-EMA codebook update strategy, where the KLD regularization prevents codebook collapse via uniform codeword distribution and the EMA updates stabilize training by balancing utilization.
Implemented via IEEE 802.11a-compliant QPSK-OFDM, our framework attains state-of-the-art reconstruction fidelity, which surpasses all benchmarks on both perceptual quality and quantitative metrics (PSNR/MS-SSIM) across dynamic channel conditions.


%




\ifCLASSOPTIONcaptionsoff
  \newpage
\fi

\end{document}